\RequirePackage{amsmath}

\documentclass{article} 

\usepackage{amsmath,amsfonts,bm}

\def\1{\bm{1}}

\def\vx{{\bm{x}}}

\DeclareMathAlphabet{\mathsfit}{\encodingdefault}{\sfdefault}{m}{sl}
\SetMathAlphabet{\mathsfit}{bold}{\encodingdefault}{\sfdefault}{bx}{n}

\usepackage[utf8]{inputenc} 
\usepackage[T1]{fontenc}    
\usepackage{hyperref}       
\usepackage{url}            
\usepackage{booktabs}       
\usepackage{amsfonts}       
\usepackage{amsmath}
\usepackage{mathtools}
\usepackage{nicefrac}       
\usepackage{microtype}      
\usepackage{todonotes}
\usepackage{subcaption}
\usepackage{multirow}
\usepackage{enumitem}
\usepackage{diagbox}
\usepackage{hyperref}
\usepackage{url}
\usepackage{bbm}
\usepackage{makecell}
\usepackage{caption}

\newcommand{\loss}{\mathcal{L}}
\graphicspath{{figs/}}

\newcommand{\wzero}{W^0}
\newcommand{\wsign}{W^+}

\newcommand{\signbit}{B_+}
\newcommand{\signbita}{\widehat{\signbit}}
\newcommand{\zerobit}{B_0}
\newcommand{\zerobita}{\widehat{\zerobit}}

\makeatletter 
\let\c@table\c@figure
\let\c@lstlisting\c@figure
\let\ftype@table\ftype@figure
\let\ftype@listings\ftype@figure
\makeatother

\usepackage[accepted]{icml2021}

\icmltitlerunning{Neural Status Registers}

\begin{document}

\twocolumn[
\icmltitle{Neural Status Registers}



\icmlsetsymbol{equal}{*}

\begin{icmlauthorlist}
\icmlauthor{Lukas Faber}{to}
\icmlauthor{Roger Wattenhofer}{to}
\end{icmlauthorlist}

\icmlaffiliation{to}{ETH Zurich}
\icmlcorrespondingauthor{Lukas Faber}{lfaber@ethz.ch}

\icmlkeywords{Machine Learning, ICML}

\vskip 0.3in
]
\printAffiliationsAndNotice{}

\begin{abstract}
Standard Neural Networks can learn mathematical operations, but they do not extrapolate. Extrapolation means that the model can apply to larger numbers, well beyond those observed during training. Recent architectures tackle arithmetic operations and can extrapolate; however, the equally important problem of quantitative reasoning remains unaddressed. In this work, we propose a novel architectural element, the Neural Status Register (NSR), for quantitative reasoning over numbers. Our NSR relaxes the discrete bit logic of physical status registers to continuous numbers and allows end-to-end learning with gradient descent. Experiments show that the NSR achieves solutions that extrapolate to numbers many orders of magnitude larger than those in the training set. We successfully train the NSR on number comparisons, piecewise discontinuous functions, counting in sequences, recurrently finding minimums, finding shortest paths in graphs, and comparing digits in images.
\end{abstract}

\section{Introduction}
Most living species are able to reason quantitatively. Ants for instance have a way to compare the size of two food sources. But not only animals, also programming languages use a lot of comparisons, e.g., in if-statements or while-loops. We believe that understanding quantitative reasoning will be helpful for a better understanding of machine intelligence.


In this paper, we propose a deep learning solution to handle quantitative reasoning. One may think that existing neural networks can already solve quantitative reasoning. 
However, this is not true. Mathematical reasoning remains a difficult domain where neural networks struggle. In particular, neural networks seem to fail to extrapolate to inputs beyond the training data. Even for simple tasks such as learning the identity function $f(x)=x$, neural networks cannot make correct predictions for numbers outside the training range~\citep{trask2018neuralalu}. Failing to extrapolate suggests that the network did not learn to reason about the actual task but rather only captured some patterns specific to the training data. This insight motivated the study of neural architectures that learn and extrapolate mathematical operations. Several works suggest architectures for arithmetical operations such as addition or multiplication~\citep{kaiser2016neural, madsen2020Neural}, however, quantitative reasoning remains unaddressed.

In this paper, we propose the Neural Status Register (NSR), a novel architectural element suited for quantitative reasoning. The NSR is inspired by the arithmetic logic unit (ALU) in modern CPUs. An ALU use status registers to allow for conditional statements. Concretely, it computes the difference between two numbers. Is the difference positive (sign bit $\signbit$) or equal to zero (zero bit $\zerobit$)? We can model all comparisons ($\leq,\neq,\ldots$) as logic combinations of these two bits. Our NSR generalizes the status registers to the continuous space. We present an NSR which allows for  end-to-end learning with gradient descent. In experiments, we show how the NSR can learn all basic quantitative reasoning tasks reliably, and how the NSR can extrapolate to numbers many \emph{orders of magnitude} larger than the training data. We show how neural networks can combine an NSR with other layers (or recurrently) to learn many interesting functions.
\begin{itemize}
    \item We present Neural Status Registers (NSRs), a novel architectural element for learnable quantitative reasoning. Our NSR is inspired by CPU status registers, which we relax to a continuous space. We perform further modifications based on theoretical insights from gradient analysis. The final NSR architecture works for both integers and floats.
    \item Unlike existing neural network architectures, the NSR is able to systematically extrapolate to make correct predictions for large numbers.
    \item The NSR enables many applications, e.g., learning piecewise discontinuous functions, counting in sequences, finding the minimum in a set, finding shortest paths in large graphs, or comparing images of digits.
\end{itemize}

\section{Related Work}
\label{sec_relatedwork}

\subsection{The Importance of Extrapolation}
Prior work discussed the importance of extrapolation to evaluate trained neural networks. \citet{santoro2018measuring} measure reasoning for several examples based on IQ tests, showing how traditional methods struggle; \citet{lake2018generalization} and \citet{suzgun2019evaluating} report the same phenomenon for language models. \citet{martius2017extrapolation} and \citet{sahoo2018learning} discuss extrapolation in the context of learning equations of control systems, for example swinging up a pendulum. They also propose a dedicated architecture for learning such equations that can extrapolate to unseen domains. \citet{xu2020neural} investigate extrapolation for feedforward and graph neural networks. Generally, they find that networks extrapolate poorly outside of the induced space of their activation function. These works show that extrapolation is ubiquitous in deep learning and that we can tackle it through dedicated architectures.

\citet{madsen2019measuring} discuss extrapolation in the context of mathematical tasks. Extrapolation can make it difficult to judge errors based on their magnitude, which is why they propose a success-based instead of an error-based quality metric. We follow this notion and measure extrapolation in terms of successes, where prediction must fall into a tight interval around the target value to be correct.

\subsection{Deep Learning for Mathematics}
\citet{kaiser2016neural} are the first to investigate neural architectures for extrapolating with their Neural GPUs. These GPUs learn addition and multiplication over binary alphabet; with being further improved by \citet{freivalds2017improving}. The Neural Arithmetic Logic Units (NALU)~\citep{trask2018neuralalu}, improved NALU~\citep{schlor2020inalu}, and the Neural Arithmetic Units (NAU)~\citep{madsen2020Neural} are two other extrapolating architectures for addition, subtraction, multiplication, and division (only the NALU). Finally, \citet{heim2020neural} extend the semantics to complex numbers, thus supporting negative inputs, division, and taking powers of inputs. However, none of these architectures support quantitative operations like the NSR. Rather, the NSR complements these architectures, for example, to allow conditional computation of functions.

In another line of research, \citet{saxton2019analysing} and \citet{lample2020deep} analyze reasoning capabilities of natural language models for mathematical tasks. These tasks include, for example, solving equations, factorizing, differential equations, or function integration. The learned models solve the training and test sets well but struggle with extrapolation in the form of larger numbers~\citep{saxton2019analysing} or differently-generated data~\citep{lample2020deep}.

\subsection{Computer-Inspired Neural Architectures}
Orthogonal work explores building differentiable versions of computer parts, most notably storage components. For example, \citet{graves2014neural}, \citet{zaremba2015reinforcement}, \citet{le2020Neural} explored the idea for having a readable and writable tape as in turing machines. \citet{weston2015memory} explore sequential memory for analysing text sequences, which \citet{sukhbaatar2015end} improve for allowing greater flexibility for reading and writing. \citet{graves2016hybrid} proposes an alternative random access memory. Finally, \citet{grefenstette2015learning} proposes differentiable data structures, such as stacks or queues.

\paragraph{Algorithm Inference.}
Plugging together computer-based differentiable architectures, we could think about assembling a differentiable computer and learn entire programs end to end with gradient descent. Some of the previous works~\citep{graves2014neural, zaremba2015reinforcement, zaremba2016learning} also learn small programs. Orthogonal work used inspiration from different programming paradigms to design architectures for algorithm inference. For example, \citet{reed2016interpreters, li2017lattice, chen2018neural} want to learn programs as compositions of simple instructions, \citet{cai2017recursion, feser2017functional} follow a functional and recursive approach, and \citet{evans2018learning, dong2019logicmachines} use logical reasoning and declarative programming.

\section{Neural Status Registers}
\begin{figure*}
    \centering
    \def\svgwidth{\textwidth}
    \input{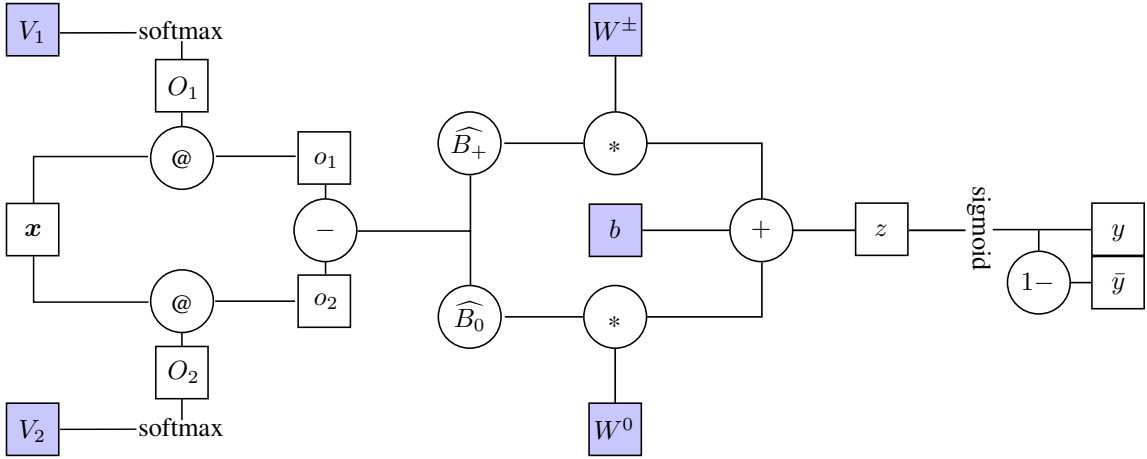}
    \caption{High-Level NSR architecture. Boxes show (intermediate) vectors and scalars and circles are operations between them. Shaded boxes indicate learnable parameters; $@$ denotes matrix multiplication. The NSR learns $V_1$ and $V_2$ to select the right elements from input vector $\vx=[x_1, x_2, \dots, x_n]$ to compare; then computes the difference between the two learnt operands. The second set of parameters $\wsign$, $\wzero$ and a bias $b$ learn to weigh $\signbit$ and $\zerobit$ activations to produce the needed comparison. The output $y$ returns whether the comparison evaluates to true or false.}
    \label{fig_nsrarchitecture}
\end{figure*}
\subsection{Architecture Overview}
The underlying idea for the NSR stems from physical circuits that employ a status register for integer comparison. These physical registers compare two numbers $x_1$ and $x_2$ through subtraction and checking two bits: a \emph{sign bit} $\signbit$ (if the difference is positive) and a \emph{zero bit} $\zerobit$ if the difference is zero. Combining these bits through logical operators (\texttt{and}, \texttt{or}, \texttt{not}), we can measure all basic comparisons: $>, <, \geq, \leq, =, \ne$. The NSR follows this idea but relaxes all the integer and bit operations to continuous space, allowing differentiability and learning with gradient descent. We show the NSR architecture in Figure~\ref{fig_nsrarchitecture}. The NSR expects a vector $x$ of numbers and compares two numbers with some comparison. The NSR learns which numbers to compare using which comparison through two sets of learnable parameters. The first pair of parameters $V_1$ and $V_2$ builds a weighted selection of the input elements in $\vx$ for the operands $o_1$ and $o_2$. The NSR computes the difference between $o_1$ and $o_2$. Based on this difference, the NSR computes relaxed versions of $\signbit$ and $\zerobit$. Replacing the logical operations from physical status registers the second set of parameters---$\wsign$, $\wzero$ and a bias term $b$--learn to combine the bit activations to realize the comparisons in an activation $y$. We show example weight allocations in Appendix~\ref{tab_nsrweights}. For easier control of downstream layers, the NSR outputs $y$ as well as its negation $\bar{y}$. We output both $y$ and $\bar{y}$, so downstream layers of the NSR have easy access to both branches of the comparison.

\subsection{Continuous Relaxation}
First, we compute derivatives for all learnable parameters. We relax notation to allow taking derivatives of vectors and assume there is only a gradient signal for $y$, not for $1-y$. We further only compute partial derivatives of $y$ with respect to the NSR components, abstracting from any downstream layers. We could get actual derivatives with the chain rule by multiplying with $\frac{\partial \loss}{\partial y}$, and all arguments remain the same. The derivatives are as follows (refer to Appendix~\ref{sec_derivatives} for their derivation):

\begin{align}
    \frac{\partial y}{\partial b} &= y(1-y)\label{eq_derivative_bias}\\
    \frac{\partial y}{\partial W^+} &= y(1-y) \signbit\label{eq_derivative_wsign}\\
    \frac{\partial y}{\partial W^0} &= y(1-y) \zerobit\label{eq_derivative_wzero}\\
    \frac{\partial y}{\partial O_1} &= y(1-y) (\signbit' W^+ + \zerobit' W^0)\label{eq_derivative_o1}\\
    \frac{\partial y}{\partial O_2} &= - y(1-y) (\signbit' W^+ + \zerobit' W^0)\label{eq_derivative_o2}
\end{align}

\begin{figure}
\centering
\includegraphics[width=0.95\columnwidth]{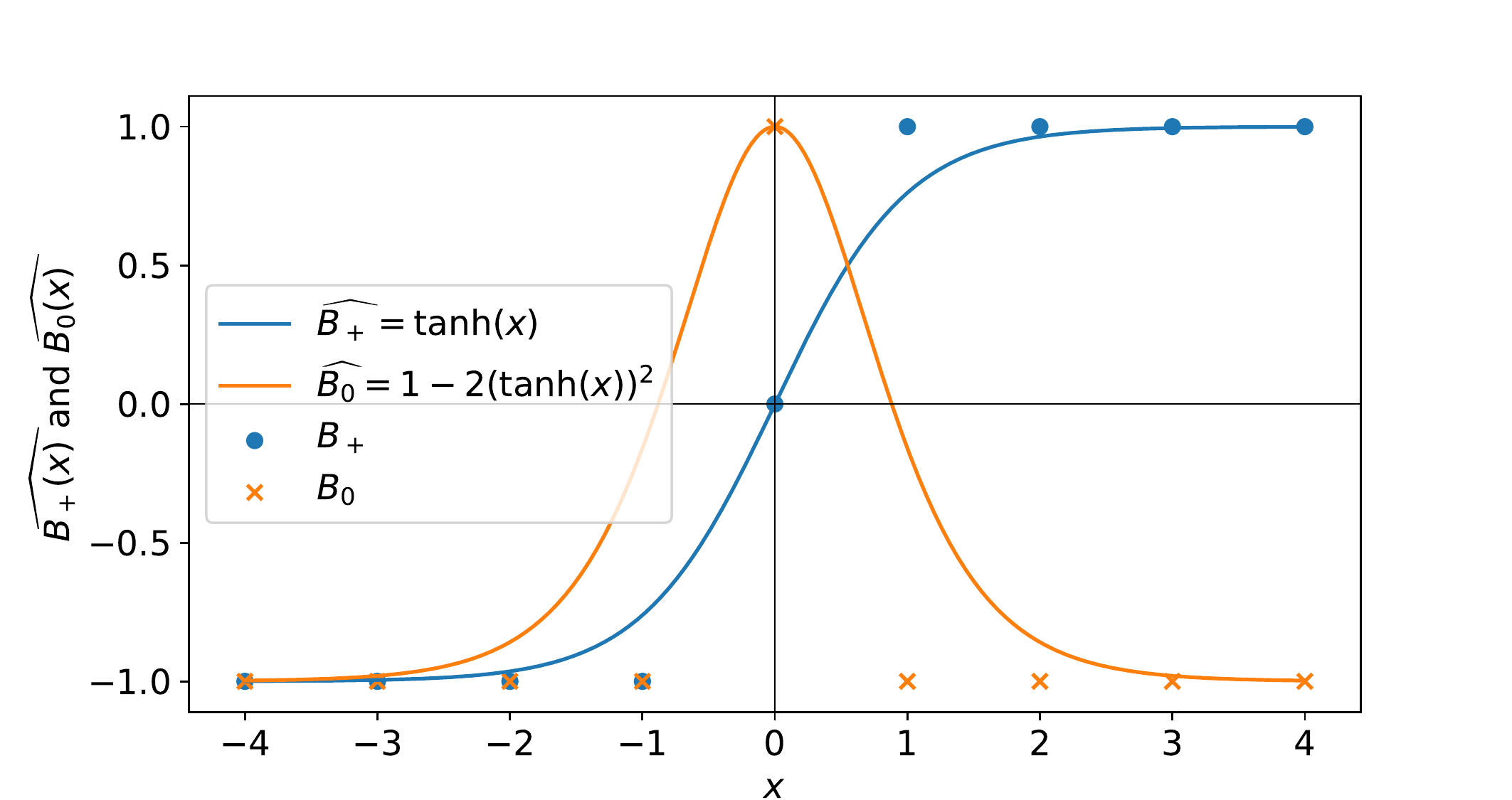}
\caption{Plots for $\signbit$ and $\zerobit$, rescalend to $[-1; 1]$ (dots and squares). They overlap for negative differences. Lines show the continuous approximations $\signbita$ and $\zerobita$.}
\label{fig_signzero}
\end{figure}

First let us address the two Equations~(\ref{eq_derivative_o1}) and \eqref{eq_derivative_o2}. To compute derivatives for $O_1$ and $O_2$ (and thus $V_1$ and $V_2$), we need to differentiate the two functions $\signbit$ and $\zerobit$. In physical status registers, these are bits defined over integers, thus having a discrete input and output. Additionally, these bits take on the value $0$ if the number is positive (for $\signbit$) or not zero (for $\zerobit$). In these cases we multiply the gradients with a $0$ in Equations~\eqref{eq_derivative_wsign} and \eqref{eq_derivative_wzero}, losing any learning signal. To prevent these $0$ bit activations from slowing down the training unnecessarily, we change the bit off value from $0$ to $1$. Furthermore, we propose the following continuous relaxations for $\signbit$ and $\zerobit$ that map well to the discrete bit values (we compare the discrete and continuous versions in Figure~\ref{fig_signzero}). Note that the approximation for $\zerobita$ looks a bit off with $\zerobita\approx-0.17$ instead of $-1$. However, experiments support that this causes no problems since $\wzero$ can easily compensate for the difference in magnitude.
\begin{align*}
    \signbita(x)&=\tanh(x)\\
    \zerobita(x)&=1-2(\tanh(x))^2
\end{align*}

\subsection{Floating-point Comparisons}
\label{sec_floats}
\begin{figure}
\centering
\includegraphics[width=0.95\columnwidth]{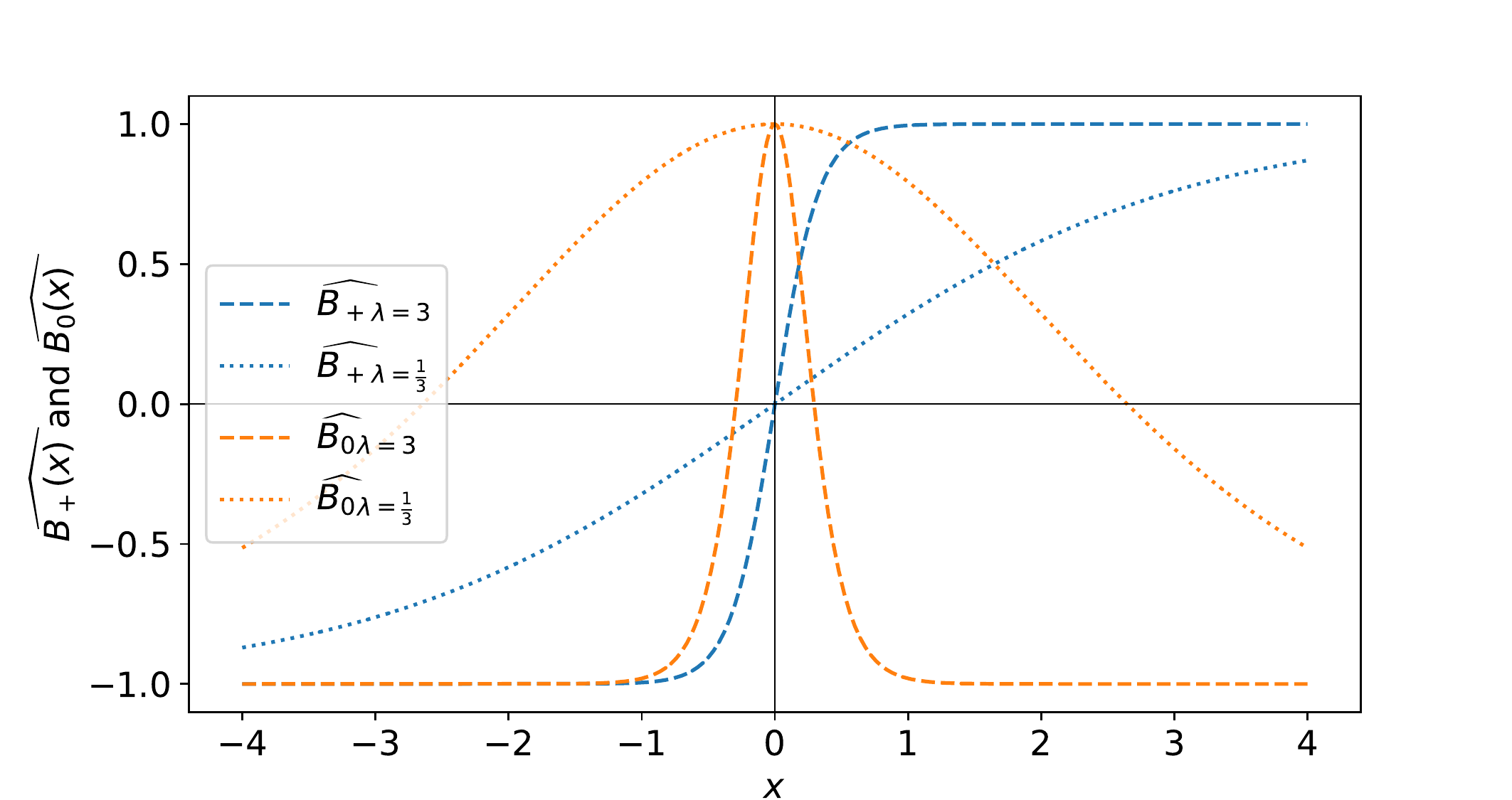}
\caption{Rescaling of approximations for the sign an zero bits. The dashed graphs show a rescaling with $\lambda=3$. Using $\lambda>1$ makes the graph more step like to produce better activations for small values of $x$. The dotted graphs use $\lambda = \frac{1}{3}$, which stretches graphs and makes $\signbita$ and $\zerobita$ saturate slower.}
\label{fig_sign_zero_scaled}
\end{figure}

As the second step, we tackle the relaxation for discrete integer inputs. As they stand in Figure~\ref{fig_signzero}, the bit relaxations support any floating numbers as inputs. However, practical learning with any inputs poses challenges. Let us look at the quality $\delta$ with which we denote the minimal difference between two non-equal numbers. In case of integers $\delta=1$. However, we would run into problems, for example if $\delta=0.5$ since $\zerobita(0.5)\approx0.57$. In this case, the sign of $\zerobita$ is incorrect, which is something that $\wzero$ cannot correct. Thus, the NSR cannot reason about (in-)equalities in that case. To a lesser extent, also $\signbita$ suffers from small delta values. For example, for $\delta\approx0$, $\signbita$ is quasi-linear and close to zero. The problem arises from the gradient Equation~\eqref{eq_derivative_wsign} having $\signbita$ as a factor, leading to slow training or even vanishing gradients. On the other hand, also $\delta>>1$ can cause problems. For a large delta, $\signbita$ and $\zerobita$ saturate to either $1$ or $-1$. In turn, their gradient approaches zero, causing vanishing gradients for the updates of $V_1$ and $V_2$ (see Equations~\eqref{eq_derivative_o1} and \eqref{eq_derivative_o2}). However, we can tackle both problems by rescaling the input to $\signbita$ and $\zerobita$ with a hyperparameter $\lambda$ before activating the bits.
\begin{align*}
    \signbita_{\lambda}(x)&= \tanh(\lambda\cdot x)\\
    \zerobita_{\lambda}(x) &= 1 - 2 \cdot \tanh(\lambda \cdot x)^2
\end{align*}
A value $\lambda>0$ makes the activation functions sharper and more ``step-like'' (see the dashed lines in Figure~\ref{fig_sign_zero_scaled}) which saturates gradients faster for large values, but alleviates problems for small values of $\delta$. On the other hand $\lambda<0$ stretches the approximations (dotted lines in Figure~\ref{fig_sign_zero_scaled}) and makes $\signbita$ and $\zerobita$ saturate later, which helps for large values of $\delta$. Generally, we can expect an inverse relationship between $\lambda$ and $\delta$. If $\delta$ increases, we can compensate with a lower value of $\lambda$ and vice versa.

\subsection{Redundancy for Reliability}
\label{subsec_lottery}
Preliminary experiments and Table~\ref{tab_compresults} show that the NSR does not always learn the comparisons $=$ and $\ne$. We hypothesize that we need the right initialization of weights to learn these function; this hypothesis is in line with other findings of the difficulty of learning XOR, which is the bit version of $\ne$~\citep{bland1998xor}. \citet{frankle2019lottery} report that while two hidden layer neurons would theoretically suffice to learn XOR, their architecture depends on lucky initialization. However, using more than two units increases the chance of learning XOR drastically. Independent of these findings \citet{kaiser2016neural} also report better robustness when they start with independent, redundant parameters that they converge towards one set of values.

Motivated by these empirical findings, we also integrate redundancy in our NSR. Instead of learning just one pair $o_1$ and $o_2$, we learn several pairs independently. For each pair, we compute the difference and activate the sign and zero bits. For each pair, we also independently learn independent values of $b$, $\wsign$, and $\wzero$. We sum all weighted bits into $z$ and proceed with a sigmoid as before.

\begin{table*}
  \caption{Learning comparisons on data from $[-10;9]$ and testing comparisons on numbers from larger orders of magnitude. For testing we sample $n$ randomly and compare $n$ against all integers at most $5$ apart; rebalanced for $=$ and $\ne$. Table cells show accuracy over $100$ runs.  MLP devolves to random guessing for $5-$ digit numbers and always predicts randomly for $=$ and $\ne$. The NSR has neither problem and solves all comparisons well.}
  \label{tab_compresults}
  \centering
  \begin{tabular}{lllllllllllllll}
    \toprule
    Task & Model & Train & $10^2$ & $10^3$ & $10^4$ & $10^5$ & $10^6$ & $10^7$ & $10^8$ & $10^9$ & $10^{10}$ & $10^{11}$ & $10^{12}$ & $10^{13}$ \\\midrule
    \multirow{6}{*}{MLP} & $>$ & \textbf{1.0} & \textbf{1.0} & 0.96 & 0.71 & 0.49 & 0.48 & 0.48 & 0.48 & 0.48 & 0.48 & 0.48 & 0.48 & 0.48\\
    & $<$ & \textbf{1.0} & \textbf{1.0} & 0.93 & 0.7 & 0.49 & 0.48 & 0.48 & 0.48 & 0.48 & 0.48 & 0.48 & 0.48 & 0.48\\
    & $\geq$ & \textbf{1.0} & \textbf{1.0} & 0.94 & 0.71 & 0.49 & 0.48 & 0.48 & 0.48 & 0.48 & 0.48 & 0.48 & 0.48 & 0.48\\
    & $\leq$ & \textbf{1.0} & \textbf{1.0} & 0.94 & 0.68 & 0.49 & 0.48 & 0.48 & 0.48 & 0.48 & 0.48 & 0.48 & 0.48 & 0.48\\
    & $=$ & 0.95 & 0.5 & 0.5 & 0.5 & 0.5 & 0.5 & 0.5 & 0.5 & 0.5 & 0.5 & 0.5 & 0.5 & 0.5\\
    & $\ne$ & 0.95 & 0.5 & 0.5 & 0.5 & 0.5 & 0.5 & 0.5 & 0.5 & 0.5 & 0.5 & 0.5 & 0.5 & 0.5\\
    \midrule
    \multirow{6}{*}{\makecell{NSR\\(ours)}} & $>$ & \textbf{1.0} & \textbf{1.0} & \textbf{1.0} & \textbf{1.0} & \textbf{1.0} & \textbf{1.0} & \textbf{1.0} & \textbf{1.0} & \textbf{1.0} & \textbf{1.0} & \textbf{1.0} & \textbf{1.0} & \textbf{1.0}\\
    & $<$ & \textbf{1.0} & \textbf{1.0} & \textbf{1.0} & \textbf{1.0} & \textbf{1.0} & \textbf{1.0} & \textbf{1.0} & \textbf{1.0} & \textbf{1.0} & \textbf{1.0} & \textbf{1.0} & \textbf{1.0} & \textbf{1.0}\\
    & $\geq$ & \textbf{1.0} & \textbf{1.0} & \textbf{1.0} & \textbf{1.0} & \textbf{1.0} & \textbf{1.0} & \textbf{1.0} & \textbf{1.0} & \textbf{1.0} & \textbf{1.0} & \textbf{1.0} & \textbf{1.0} & \textbf{1.0}\\
    & $\leq$ & \textbf{1.0} & \textbf{1.0} & \textbf{1.0} & \textbf{1.0} & \textbf{1.0} & \textbf{1.0} & \textbf{1.0} & \textbf{1.0} & \textbf{1.0} & \textbf{1.0} & \textbf{1.0} & \textbf{1.0} & \textbf{1.0}\\
    & $=$ & \textbf{0.99} & \textbf{0.93} & \textbf{0.93} & \textbf{0.93} & \textbf{0.93} & \textbf{0.93} & \textbf{0.93} & \textbf{0.93} & \textbf{0.93} & \textbf{0.93} & \textbf{0.93} & \textbf{0.93} & \textbf{0.93}\\
    & $\ne$ & \textbf{0.99} & \textbf{0.92} & \textbf{0.92} & \textbf{0.92} & \textbf{0.92} & \textbf{0.92} & \textbf{0.92} & \textbf{0.92} & \textbf{0.92} & \textbf{0.92} & \textbf{0.92} & \textbf{0.92} & \textbf{0.92}\\
    \bottomrule
  \end{tabular}
\end{table*}
\section{Experiments on the NSR}

\subsection{Learning Comparisons}
In this first set of experiments, we validate the NSR for the core taks of comparison. We compare the NSR against a standard feedforward network (MLP). We use an MLP with one hidden layer and one output neuron, using sigmoid activations. We use a hidden layer dimension of $20$, which gives the model $81$ parameters. In comparison, an NSR with redundancy of $10$ also has $81$ parameters, allowing for a fair comparison. We show how the MLP can theoretically learn all comparisons in Appendix~\ref{tab_mlpweights}.

We train on all integer pairs from $[-10; 9]$ and run experiments for all comparisons $>, <, \geq, \leq, =, \ne$. The label for a pair of numbers is $1$ if the comparison is true, $0$ otherwise. As a consequence, $=$ and $\ne$ have imbalanced training sets. We supervise using the Mean Absolute Error and train for $50.000$ epochs. We use the Adam~\citep{kingma2015adam} optimizer with its default settings from the PyTorch~\citep{paszke2015pytorch} library, including the learning rate.

After training, we evaluate if the learned model can extrapolate to unseen data. We follow the powers of ten for extrapolation checks. For every check, we sample a random number $n$ is in the same order of magnitude as the column header. Then, we create a test set by comparing $n$ against every other close-by integer in the range $[n-5;n+5]$, including itself for a total of $11$ tests. For $=$ and $\ne$ these sets are imbalanced, therefore we add test cases ($n+i, n+i)$ with $i \in [-5;5]\setminus\{0\}$. We measure the comparison accuracy for this test set. Table~\ref{tab_compresults} reports the accuracy averaged over $100$ runs. We can see NSR achieves excellent extrapolation across all comparisons while MLP degenerates to random guessing with $5+$ digits. The NSR also reliably learns the comparisons $=$ and $\ne$, which MLP does not learn at all (the $0.95$ accuracy during training stems from constantly predicting the majority answer in the unbalanced training set). Apart from the imbalanced training set, these two comparisons also have a harder loss landscape which makes learning dependent on the right initialization~\citep{bland1998xor}.

\subsection{Learning with Floats}
\begin{figure}
    \centering
    \includegraphics[width=\columnwidth]{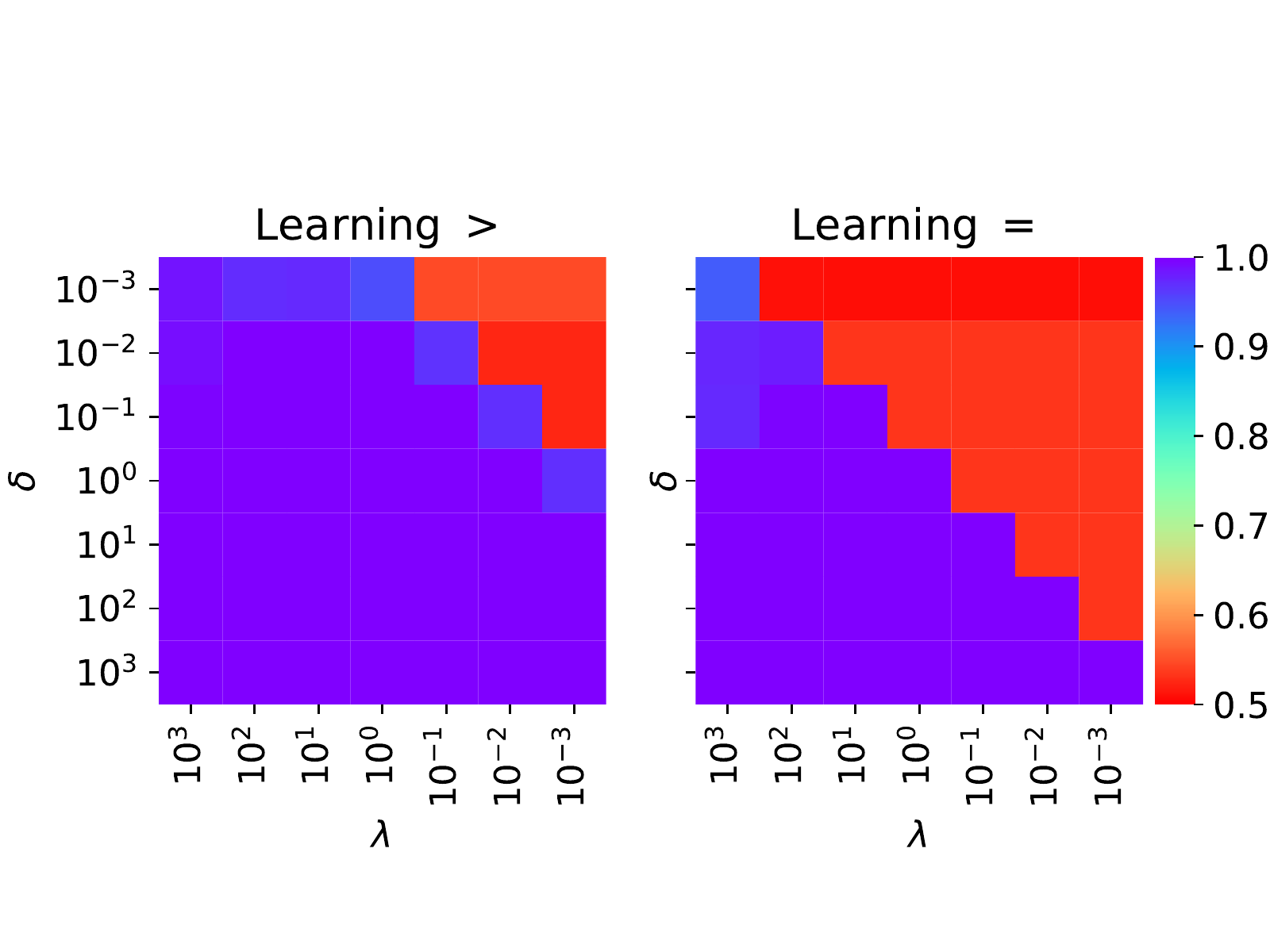}
    \caption{Learning binary comparisons with the NSR for floats. $\delta$ denotes the minimum distance between two non-equal numbers and $\lambda$ the scale factor for $\signbita$ and $\zerobita$ logits. The training and evaluation setup is the same as in Table~\ref{tab_compresults}. $\lambda$ compensates for $\delta$ values different from $1$ (integers); if $\lambda\geq\delta^{-1}$ for learning $=$ or $\lambda\geq10^3\delta^{-1}$ for learning $>$, the NSR can train successful. Accuracy averaged over $20$ runs.}
    \label{fig_floats}
\end{figure}

In this experiment, we investigate reasoning over floats. As discussed in Section~\ref{sec_floats}, the important measure is the minimum difference $\delta$ between two non-equal values. We can compensate for $\delta$ with scaling the pre-activation values for $\signbita$ and $\zerobita$ with a scalar $\lambda$. To investigate the impact, we run different experiments, varying both $\delta$ and $\lambda$ across $\{10^{-3}, 10^{-2}, 10^{-1}, 1, 10^1, 10^2, 10^3\}$. We multiply every input in the training set---coming again from $[-10; 9]$---with $\delta$ to obtain inputs with a smaller difference. For testing, we also scale the test numbers accordingly to be only $\delta$ apart from each other. The remaining setup is as in the previous section. For every pair $\delta,\lambda$, we report $>$ as an easy comparison, and $=$ as a difficult comparison, the results extend to the other comparisons (see Appendix~\ref{sec_allfloats}). Due to the clear results, we only perform $20$ instead of $100$ runs. For every pair $(\delta; \lambda)$ we get a series of extrapolation results, equivalent to one row in Table~\ref{tab_compresults}. In Figure~\ref{fig_floats}, we show the average of all these extrapolation tests, for each pair $(\delta; \lambda)$.

For both comparisons, there is a threshold where the NSR learns the comparison almost perfectly and or not learning at all. If both $\delta$ and $\lambda$ are small, $\signbita$ becomes small enough to cause a vanishing gradient problem. Then the NSR cannot learn $>$. The NSR can learn $=$ if $\lambda\geq\delta^{-1}$. Since in these cases the sign of $\signbita{\delta)}$ remains negative.
We take away that the proposed scaling through $\lambda$ works. While $\lambda$ is a hyperparameter, we saw that a good rule of thumb is to set $\lambda$ to the inverse of the assumed smallest difference $\delta$ in the expected data distribution. Moreover, the NSR is forgiving for having a larger value of $\lambda$, which gives some room for safety and error. We can increase $\lambda$ few orders of magnitude higher than our expected $\delta^{-1}$. Thus, we conclude that the NSR can handle floats as well as integers. For the remainder of the experiments, we will consider integers for simplicity.

\begin{table*}
  \caption{Learning piecewiese defined functions. The numbers in the ``Train'' and following columns (showing extrapolation) denote the accuracy for prediction averaged over $100$ runs. Predictions are made based on comparisons of a $5-$ interval around a random number $p$. The number is randomly drawn from the order of magnitude given in the column header. An accurate prediction is off by at most $0.1$ from the true target.}
  \label{tab_functions}
  \centering
  \begin{tabular}{lllllllllllllll}
    \toprule
    Model & Function & Train & $3^2$ & $3^3$ & $3^4$ & $3^5$ & $3^6$ & $3^7$ & $3^8$ & $3^9$ & $3^10$ & $3^{11}$ & $3^{12}$ & $3^{13}$\\\midrule
    \multirow{2}{*}{MLP} & abs & \textbf{1.0} & \textbf{1.0} & 0.98 & 0.57 & 0.01 & 0.01 & 0.0 & 0.0 & 0.0 & 0.0 & 0.0 & 0.0 & 0.0\\
    & f & \textbf{1.0} & \textbf{1.0} & \textbf{1.0} & 0.99 & 0.95 & 0.85 & 0.7 & 0.55 & 0.47 & 0.4 & 0.31 & 0.2 & 0.09\\
    \midrule
    \multirow{2}{*}{NSR} & abs & \textbf{1.0} & 0.99 & \textbf{0.99} & \textbf{0.99} & \textbf{0.99} & \textbf{0.99} & \textbf{0.99} & \textbf{0.99} & \textbf{0.99} & \textbf{0.99} & \textbf{0.99} & \textbf{0.99} & \textbf{0.99}\\
    & f & 0.99 & \textbf{1.0} & \textbf{1.0} & \textbf{1.0} & \textbf{0.99} & \textbf{0.96} & \textbf{0.87} & \textbf{0.69} & \textbf{0.52} & \textbf{0.42} & \textbf{0.38} & \textbf{0.36} & \textbf{0.36}\\
  \end{tabular}
\end{table*}
\subsection{Learning Downstream Units: Piecewise Functions}
We now show that we can combine the quantitative reasoning of the NSR with Neural Arithmetic Units to learn extrapolating models for piecewise defined functions. As an example, we look at the following two functions: 
\begin{align}
    abs(x_1, x_2) = \begin{cases}x_1 - x_2 & \text{ if } x_1 > x_2 \\ x_2 - x_1 & \text{ else} \end{cases}\nonumber\\
    f(x_1,x_2,x_3,x_4,x_5) =\begin{cases}x_5 + 4 & \text{ if }x_1 > x_2\\ x_4 - x_3 & \text{ else}\end{cases}\nonumber
\end{align}
We expect the first function $abs$ to be easy to learn since it only depends on two inputs and is still a continuous function. On the other hand, $f$ has five inputs and is not continuous, which is why we expect that $f$ is harder to learn.

The training set contains one data point for every pair from $[-10; 9]$ for the two numbers in the comparisons. We sample the remaining inputs from $[-100; 100]$. We again evaluate trained models on their extrapolation to higher numbers. The extrapolation is not as good as in the example of learning comparisons. Therefore, we look at orders of magnitude based on the power series of $3$. We repeat the same test setting as before, we draw a random number in the given order of magnitude, and set up test cases with all numbers in the $5-$ interval around for the comparison inputs. We sample the remaining inputs from $[-100; 100]$. We only consider predictions on these test sets correct if the absolute deviation from the true value is st most $0.1$. We report the accuracy values, averaged over $100$ runs in Table~\ref{tab_functions}.

Again, the NSR shows superior extrapolation. The NSR is clearly better for learning $abs$, where MLP effectively does not extrapolate beyond scaling with $3^4$. For $f$, we see the NSR being consistently better across all extrapolation tests, however, to a lesser extent. Here both models suffer from numerical problems that come with functions such as sigmoid saturating. When facing large numbers, such as $3^8$, even a strongly saturated sigmoid activation of $0.999$ will propagate errors when we multiply with it. We leave a principled solution to this for future study.

\subsection{Redundancy Ablation}
\begin{figure}
    \centering
    \includegraphics[width=\columnwidth]{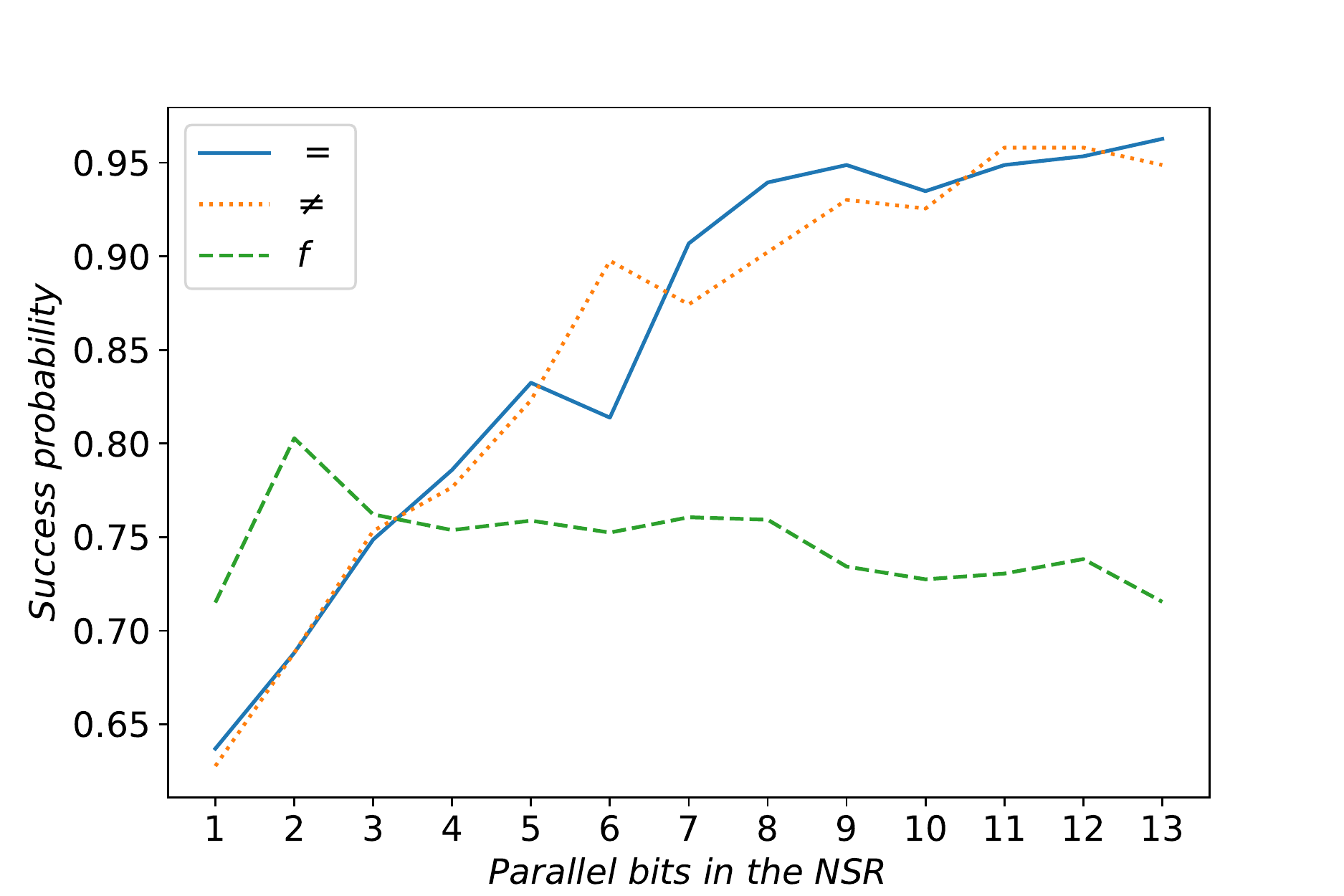}
    \caption{Average accuracy over extrapolation checks ($y-$axis), for varying redundancy in the NSR ($x-$axis). Higher redundancy helps in learning the tasks better but the effect starts to plateau at around $9$, then start slight deteriorating for one task.}
    \label{fig_lottery}
\end{figure}

Next, we investigate an ablation on the redundancy from Section~\ref{subsec_lottery}. Here, we explore different previous problems: $=$, $\ne$, $f$. We vary the redundancy from $1$ to $15$; for each redundancy, we run one experiment and use the same settings for each task as in the previous settings. We plot the accuracy overall extrapolation checks in Figure ~\ref{fig_lottery}.
The two comparisons $=$ and $\ne$ benefit greatly from redundancy. Their accuracy scores keep increasing until around $9$ redundant units. However, these tasks do not benefit from more redundancy. On the other hand, $f$ does not benefit from redundancy, nor does redundancy hurt. There seems to be a peak at redundancy $2$ and a slight decline with increasing redundancy, but neither effect is strong. We take away that redundancy generally helps (or at least does not hurt). Having a redundancy of $10$ as used in the previous experiments appears to be a reasonable value we continue to use in subsequent experiments.

\subsection{The Recurrent NSR}
\begin{figure}
    \centering
    \includegraphics[width=0.95\columnwidth]{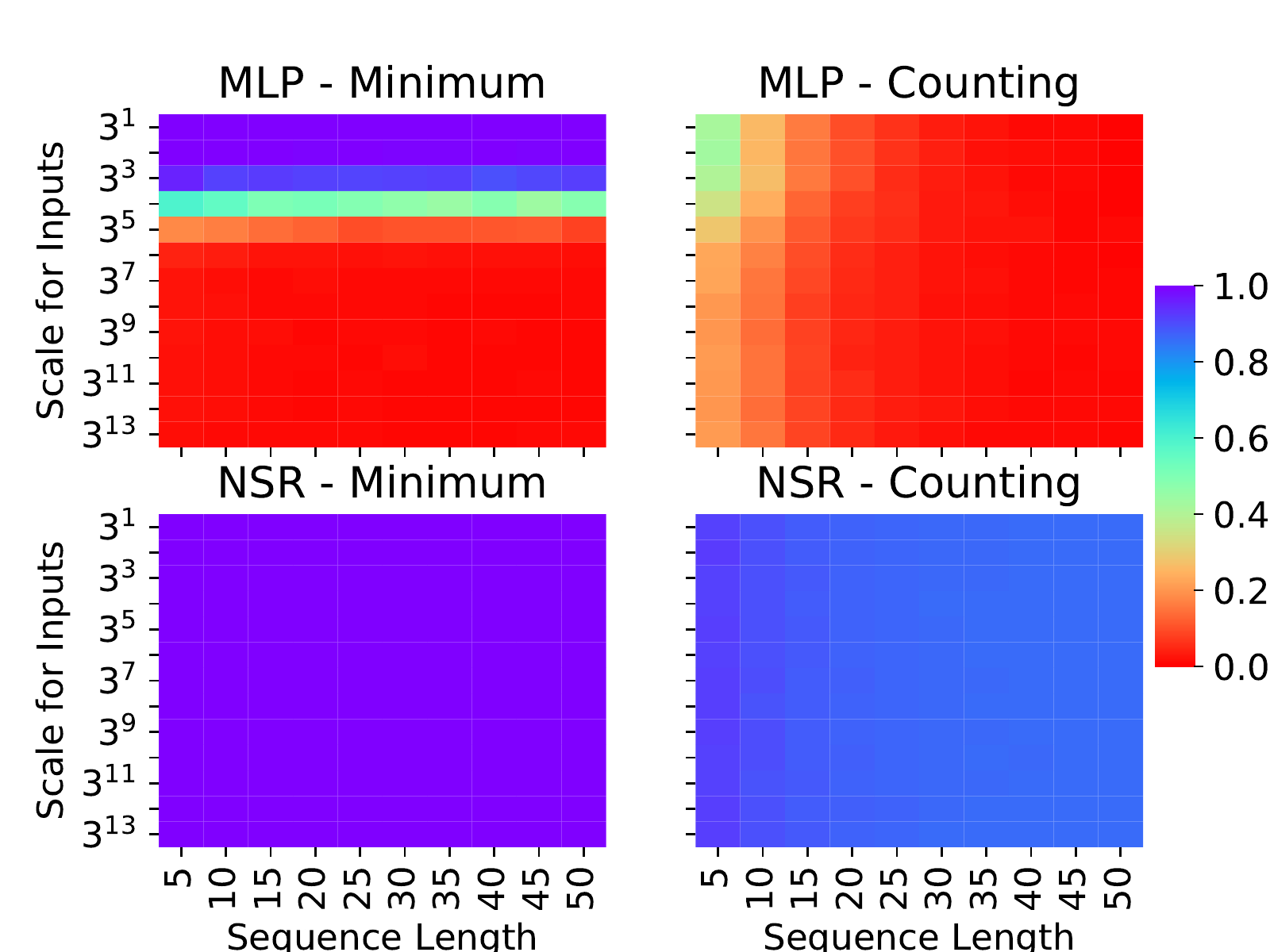}
    \caption{Learning recurrent problems with MLP and the NSR. The left plots show learning the minimum of a sequence, right plots to count the number of occurrences of an element in a sequence. Higher numbers are better. The NSR learns both tasks and scales both to higher sequence numbers and to more sequence elements.}
    \label{fig_recurrent}
\end{figure}

In this further experiment, we use the NSR as a recurrent layer. We learn the minimum in a list and learn how often a particular number occurs in the rest of the sequence. For learning the minimum, the NSR has one state cell that it can use to keep track of its belief of the minimum. In one step, it can compare this cell to the next input and update its state as a linear combination of the two. For learning to count, one cell stores the first element of a list, another the counter. In one step, the NSR receives the next element of the list and can learn to control an increment unit manipulating the counter. For comparison we use the same setup with a standard RNN architecture with an MLP as the recurrent layer. Note that using recurrent layers such as the GRU or LSTM will not produce good results, because they non-linearly squash their inputs in every step. Thus they, cannot reproduce the final numbers which are linearly dependent on the inputs, which is why we did not test these architectures.
We train on $500$ sample lists of length $5$ where elements can be any number $\in [-10;9]$. We extrapolate along two dimensions for testing: we try lists of increasing lengths by increments of $5$, up to $50$. As before, we also extrapolate the number range for comparison by sampling from the orders of magnitude defined by $3^i$ and use numbers from the $5-$interval around the sampled number. We test $50$ such samples and consider one sample correct if the prediction is at most $0.1$ percent off. Figure~\ref{fig_recurrent} shows the accuracy averaged over $100$ runs.
MLP does not scale to higher numbers when learning the minimum, stopping extrapolation starting from a factor of $3^4$. However, it can scale across the sequence length. For learning to count (which requires learning $=$), MLP fails altogether, similar to learning only $=$ in Table~\ref{tab_compresults}. On the other hand, the NSR learns the minimum perfectly across all sequence lengths and extrapolation scales. The NSR also learns counting well, independent of the extrapolation scale. However, the NSR shows slight deterioration when increasing the sequence length, going down to around $86\%$ accuracy.

\subsection{Shortest Paths in Graphs}
\begin{figure}
    \centering
    \includegraphics[width=0.90\columnwidth]{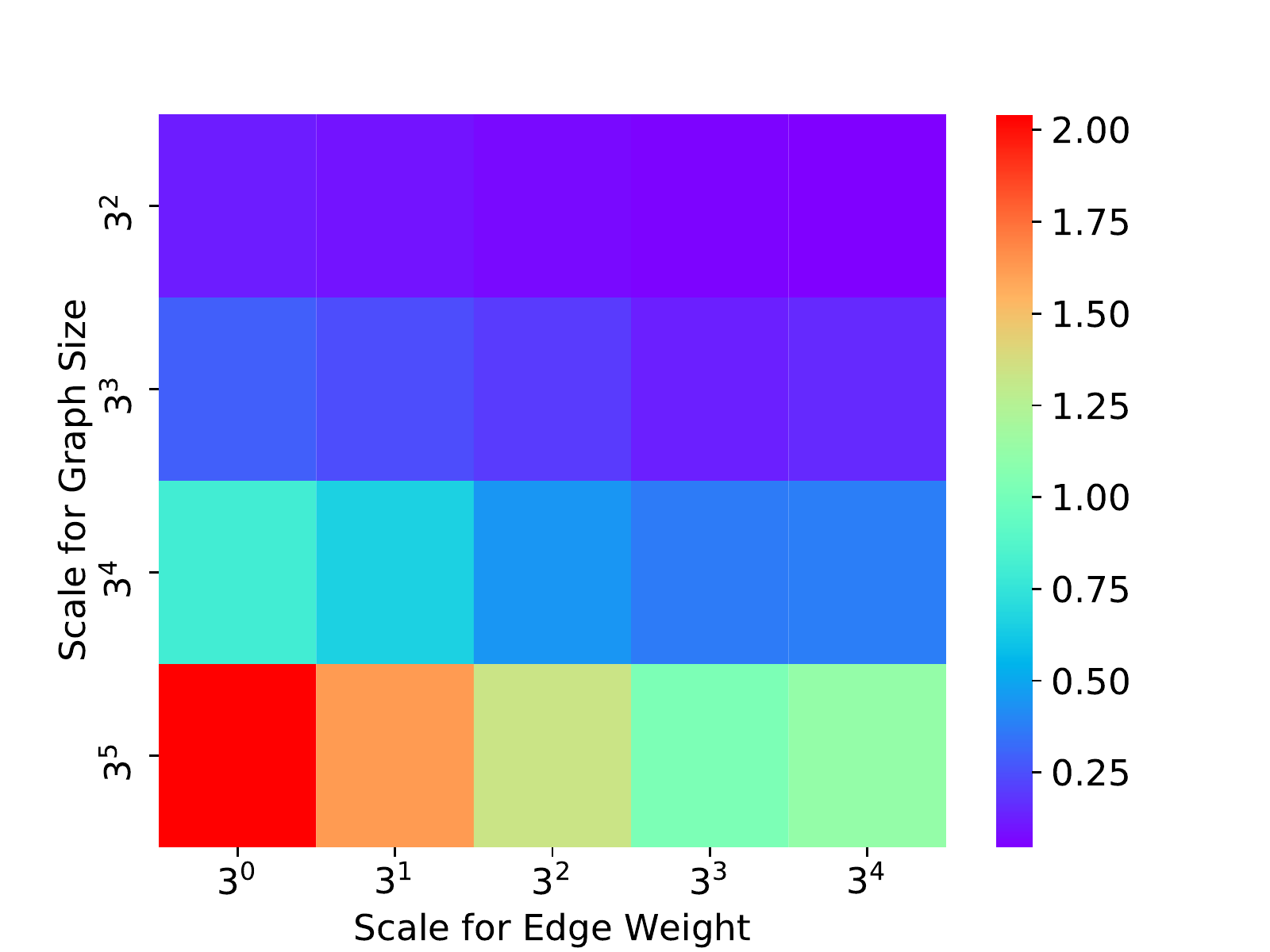}
    \caption{Learning shortest paths with a graph neural network, where the NSR learns the aggregate function. Higher numbers are better. The NSR learns the minimum successfully and extrapolates across more nodes and higher edge weights.}
    \label{fig_sssp}
\end{figure}

As a follow-up from the encouraging results using the recurrent NSR to find the minimum, we now use it to search for shortest paths with a message passing Graph Neural Network (GNN~\citep{gilmer2017neural}). Generally, we follow the idea from~\citet{velickovic2020neural, tang2020towards}; we have an input graph with weighted edges and want to learn the shortest path distances from a source node. To do this, we want the network to learn to imitate the Bellman-Ford algorithm. In every step, a node in the graph receives its current distance and all its neighbors' distances plus the length of the edges connecting to them. The new distance is the minimum of all these values. This experiment shows that the previous experiment's recurrent NSR can learn to aggregate the numbers to a minimum. The previous works hardcoded this knowledge to take the minimum by using a $max$~\citep{velickovic2020neural} or $min$~\citep{tang2020towards} in their graph neural network for aggregating the information (also called ``messages'') from their neighbors. We generate $25$ graphs, each having $n=10$ nodes with a spanning tree base structure. We uniformly add further edges with probability $frac{1}{2n}$. As in \citet{velickovic2020neural}, we supervise with the Bellman-Ford update equation using teacher forcing for intermediate steps. We leave out their termination network (running $n$ iterations) and messaging networks (using the correct message of node state plus edge weight) to focus on evaluating the NSR. After training, we expose the trained model to extrapolated inputs, where we scale both the maximum edge weight and the number of nodes with $3^i$, and sample from that order of magnitude. We measure performance as the mean average error from the predicted distance to the ground-truth distance computed by Bellman-Ford, normalized by the maximum possible weight. Figure~\ref{fig_sssp} shows the results across $10$ random test graphs per seed, for $10$ seeds.

The NSR learns to aggregate the neighbor messages with the minimum reaching close to $0$ error. The trained models convincingly scale to both larger graphs and larger graph weights. When we increase the nodes by an average factor of $3$---and thus have $3$ times as many iterations, the errors grow slower. Interestingly, the NSR performs better instead of worse when we increase the edge weights. This is because both larger weights larger shortest paths and with greater distances between numbers. The NSR can capitalize on those large distances to produce better results despite allegedly more difficult inputs. For future work, we want to study this potential in more detail, especially in conjunction with IterGNNs~\citep{tang2020towards}, which could complement the NSR by learning those components we hard coded in our example (termination and the messaging function).

\subsection{Learning Upstream Units: MNIST}
\begin{figure}
    \centering
    \includegraphics[width=0.9\columnwidth]{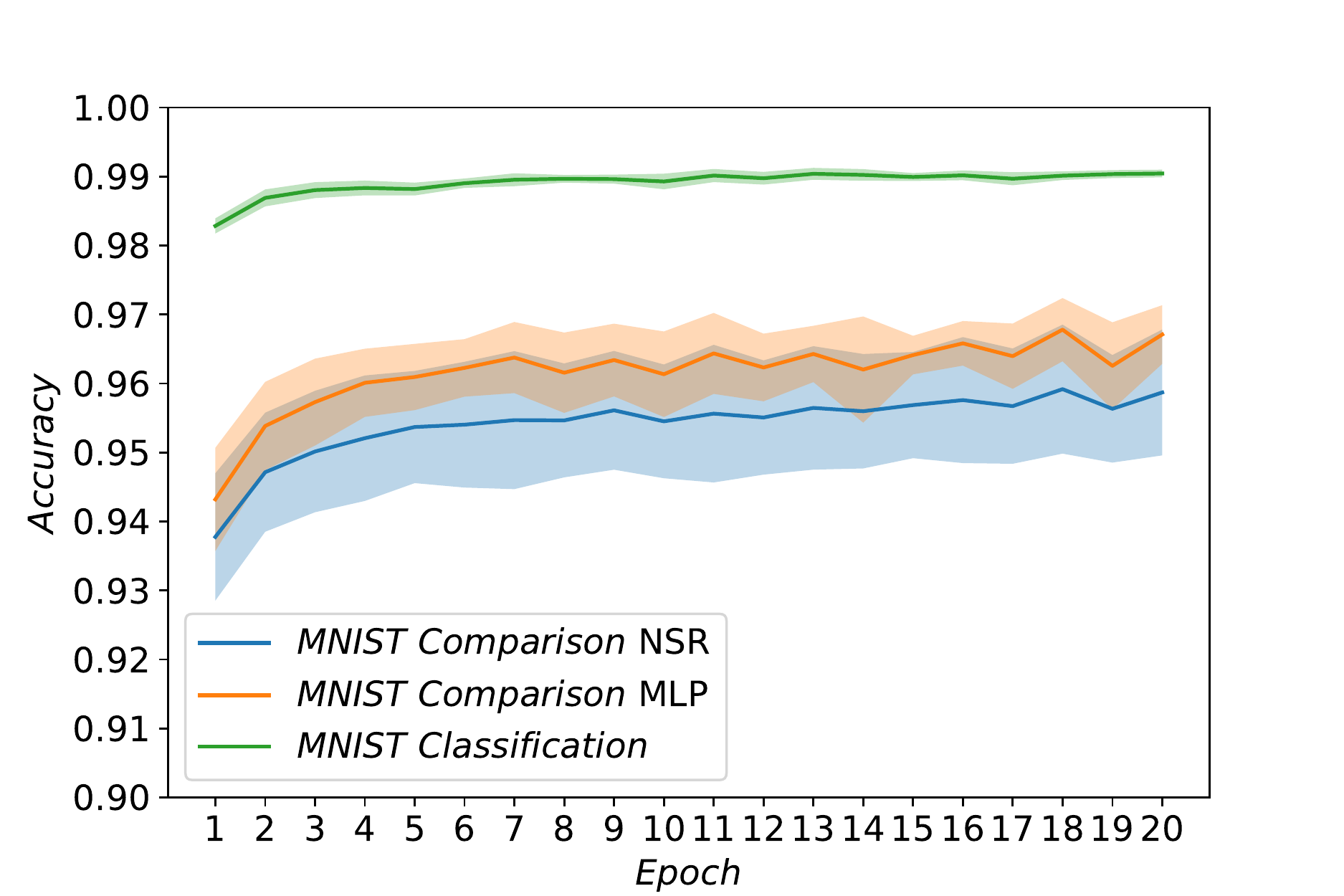}
    \caption{Accuracy (y-axis) for digit comparison task for NSR (blue) and MLP (orange), over training for multiple epochs (x-axis). Digit classification for reference(green).}
    \label{fig_mnist}
\end{figure}

Finally, we employ the NSR as a downstream unit for a convolutional neural network and show we can learn the entire network end-to-end to solve a digit comparison task. We input two MNIST pictures of digits and the model has to output $1$ if the first digit is larger than the second, otherwise $0$. We use a small convolutional neural network for digit classification, where we read out class probabilities and multiply with the row vector of digits $[0;1;2;3;4;5;6;7;8;9]$. This predicts the digit in the image, which we feed in a comparison model (NSR or an MLP). We supervise the entire model using only the comparison's supervision, never revealing the actual digit in an image. We split the MNIST training set into batches of $50$ images for our task, and one training batch is all pairs of images per image batch. We run without learning rate decay and average over $10$ runs. Figure~\ref{fig_mnist} shows the accuracy on the MNIST test set, also batched in groups of $50$ images. The shaded region is one standard deviation. For reference, we plot the test accuracy for the vanilla digit-recognition task using only the CNN.

The MNIST comparison task seems more generally more difficult than the digit classification task. Both the NSR and MLP version do not reach the classification accuracy of the vanilla MNIST task (plotted as a reference). We expected this result since the learning signal is much weaker and less consistent. In digit recognition, an image showing a $4$ always receives the same supervision. In the comparison tasks, even only finding out the digit for an image is much harder because the same image should sometimes produce a $0$ and sometimes a $1$ in the output. Thus, we need to compare an image against at least one image of every digit to identify the digit. But the network neither knows the label for these other images, which makes this task so difficult. Comparing both comparison architectures, the NSR is on par with MLP. This is a success, since it shows the NSR can train previous layers as good as feedforward layers. However, this task does not have any extrapolation where the NSR might outperform MLP.

\section{Discussion and Conclusion}
In this paper, we introduced the Neural Status Registers (NSR). The NSR is a novel layer for quantitative reasoning on numbers, inspired from physical circuits with adaptions for end-to-end differentiability. The NSR shows extrapolation to numbers many orders of magnitude larger, leading to believe it understands the true reasoning required for comparing numbers. We can combine the NSR with other layers to learn piecewise discontinuous functions, use the NSR recurrently to count and learn the minimum, learn shortest paths, or use it with a CNN to compare digits in images. Furthermore, The NSR is robust to the choice of hyperparameters, most notably the in-built redundancy and the minimum distance of non-equal numbers.

Going forward, we want further explore the NSR for algorithm inference. Here, we see potential for algorithm inference with little supervision by combining the NSR with other recent developments, such as the NAU and IterGNNs.

\bibliography{paper}
\bibliographystyle{icml2021}

\appendix

~\newpage
\section{Derivatives}
\label{sec_derivatives}
Here we provide the derivation for the gradients of the learnable parameters with respect to $y$. The gradient for $z$ is the gradient of the sigmoid function.
\begin{align*}
    \frac{\partial y}{\partial z} &= y(1-y)
\end{align*}
With this we can compute the gradients for $\signbita$ and $\zerobita$ (which we need for $V_1$ and $V_2$) and $\wsign$, $\wzero$, and $b$ by backpropagating the summation.
\begin{align*}
    \frac{\partial y}{\partial b} &= \frac{\partial y}{\partial z}\cdot \frac{\partial z}{\partial b} &= y(1-y)\\
    \frac{\partial y}{\partial \wsign} &= \frac{\partial y}{\partial z}\cdot \frac{\partial z}{\partial \wsign} &= y(1-y)\signbita\\
    \frac{\partial y}{\partial \wzero} &= \frac{\partial y}{\partial z}\cdot \frac{\partial z}{\partial \wzero} &= y(1-y)\zerobita\\
    \frac{\partial y}{\partial \signbita} &= \frac{\partial y}{\partial z}\cdot \frac{\partial z}{\partial \signbita} &= y(1-y)\wsign\\
    \frac{\partial y}{\partial \zerobit} &= \frac{\partial y}{\partial z}\cdot \frac{\partial z}{\partial \zerobita} &= y(1-y)\wzero\\
\end{align*}
From here we get derivatives for $o_1$ and $o_2$. The two gradients are identical, bar their sign.
\begin{align*}
    \frac{\partial y}{\partial o_1} &= \frac{\partial y}{\partial o_1}\cdot \frac{\partial o_1}{\partial \signbita} + \frac{\partial y}{\partial o_1}\cdot \frac{\partial o_1}{\partial \zerobita}\\&= y(1-y)(\signbita'\wsign + \zerobita'\wzero)\\
    \frac{\partial y}{\partial o_2} &= -y(1-y)(\signbita'\wsign + \zerobita'\wzero)\\
\end{align*}
Principally, we would need to backpropagate through the matrix multiplication with $x$ and through the softmax to obtain the derivatives for $V_1$ and $V_2$. For the scope of this paper, it was important to highlight that $\signbita'$ and $\zerobita'$ are part of these gradients and thus need to be differentiable. We can conclude this from the derivatives for $o_1$ and $o_2$ and thus stop here.

\newpage

\section{Recurrrent Architectures}

\begin{figure}[!h]
    \centering
    \def\svgwidth{0.6\columnwidth}
\begingroup%
  \makeatletter%
  \providecommand\color[2][]{%
    \errmessage{(Inkscape) Color is used for the text in Inkscape, but the package 'color.sty' is not loaded}%
    \renewcommand\color[2][]{}%
  }%
  \providecommand\transparent[1]{%
    \errmessage{(Inkscape) Transparency is used (non-zero) for the text in Inkscape, but the package 'transparent.sty' is not loaded}%
    \renewcommand\transparent[1]{}%
  }%
  \providecommand\rotatebox[2]{#2}%
  \newcommand*\fsize{\dimexpr\f@size pt\relax}%
  \newcommand*\lineheight[1]{\fontsize{\fsize}{#1\fsize}\selectfont}%
  \ifx\svgwidth\undefined%
    \setlength{\unitlength}{172.13176218bp}%
    \ifx\svgscale\undefined%
      \relax%
    \else%
      \setlength{\unitlength}{\unitlength * \real{\svgscale}}%
    \fi%
  \else%
    \setlength{\unitlength}{\svgwidth}%
  \fi%
  \global\let\svgwidth\undefined%
  \global\let\svgscale\undefined%
  \makeatother%
  \begin{picture}(1,0.67621056)%
    \lineheight{1}%
    \setlength\tabcolsep{0pt}%
    \put(0,0){\includegraphics[width=\unitlength,page=1]{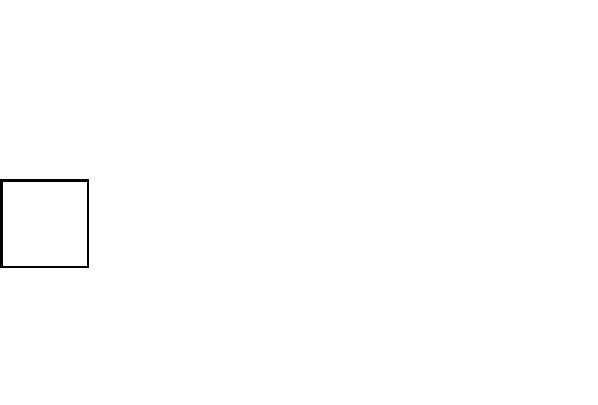}}%
    \put(0.0385,0.285){\color[rgb]{0,0,0}\makebox(0,0)[lt]{\lineheight{1.25}\smash{\begin{tabular}[t]{l}$m_i$\end{tabular}}}}%
    \put(0,0){\includegraphics[width=\unitlength,page=2]{rmin.pdf}}%
    \put(0.042,0.055){\color[rgb]{0,0,0}\makebox(0,0)[lt]{\lineheight{1.25}\smash{\begin{tabular}[t]{l}$x_i$\end{tabular}}}}%
    \put(0,0){\includegraphics[width=\unitlength,page=3]{rmin.pdf}}%
    \put(0.678,0.543){\color[rgb]{0,0,0}\makebox(0,0)[lt]{\lineheight{1.25}\smash{\begin{tabular}[t]{l}@\end{tabular}}}}%
    \put(0,0){\includegraphics[width=\unitlength,page=4]{rmin.pdf}}%
    \put(0.645,0.3){\color[rgb]{0,0,0}\makebox(0,0)[lt]{\lineheight{1.25}\smash{\begin{tabular}[t]{l}NSR\end{tabular}}}}%
    \put(0.28,0.59){\color[rgb]{0,0,0}\makebox(0,0)[lt]{\lineheight{1.25}\smash{\begin{tabular}[t]{l}$m_{i+1}$\end{tabular}}}}%
  \end{picture}%
\endgroup%

    \caption{Recurrent Minimum Finding with the NSR. The current estimate of the minimum $m_i$ is compared against the $i-$th input $x_i$ with the NSR. The NSR output controls if the new minimum $m_i+1$ is the old minimum $m_i$ or $x_i$. For the MLP experiments, we replace the NSR through an MLP.}
    \label{fig_rmin}
\end{figure}

\begin{figure}[!h]
    \centering
    \def\svgwidth{0.6\columnwidth}
\begingroup%
  \makeatletter%
  \providecommand\color[2][]{%
    \errmessage{(Inkscape) Color is used for the text in Inkscape, but the package 'color.sty' is not loaded}%
    \renewcommand\color[2][]{}%
  }%
  \providecommand\transparent[1]{%
    \errmessage{(Inkscape) Transparency is used (non-zero) for the text in Inkscape, but the package 'transparent.sty' is not loaded}%
    \renewcommand\transparent[1]{}%
  }%
  \providecommand\rotatebox[2]{#2}%
  \newcommand*\fsize{\dimexpr\f@size pt\relax}%
  \newcommand*\lineheight[1]{\fontsize{\fsize}{#1\fsize}\selectfont}%
  \ifx\svgwidth\undefined%
    \setlength{\unitlength}{166.64178449bp}%
    \ifx\svgscale\undefined%
      \relax%
    \else%
      \setlength{\unitlength}{\unitlength * \real{\svgscale}}%
    \fi%
  \else%
    \setlength{\unitlength}{\svgwidth}%
  \fi%
  \global\let\svgwidth\undefined%
  \global\let\svgscale\undefined%
  \makeatother%
  \begin{picture}(1,0.67985389)%
    \lineheight{1}%
    \setlength\tabcolsep{0pt}%
    \put(0,0){\includegraphics[width=\unitlength,page=1]{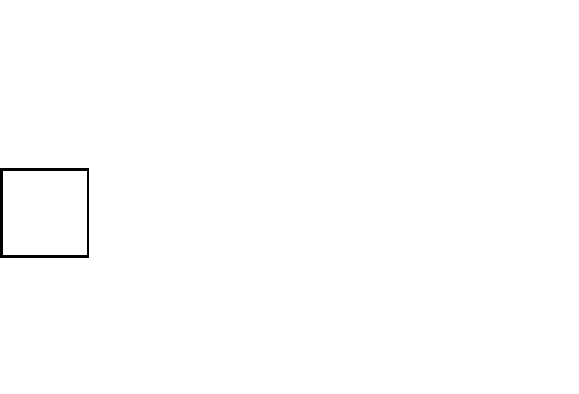}}%
    \put(0.0412,0.298){\color[rgb]{0,0,0}\makebox(0,0)[lt]{\lineheight{1.25}\smash{\begin{tabular}[t]{l}$x_0$\end{tabular}}}}%
    \put(0,0){\includegraphics[width=\unitlength,page=2]{rcount.pdf}}%
    \put(0.0412,0.063){\color[rgb]{0,0,0}\makebox(0,0)[lt]{\lineheight{1.25}\smash{\begin{tabular}[t]{l}$x_i$\end{tabular}}}}%
    \put(0,0){\includegraphics[width=\unitlength,page=3]{rcount.pdf}}%
    \put(0.675,0.568){\color[rgb]{0,0,0}\makebox(0,0)[lt]{\lineheight{1.25}\smash{\begin{tabular}[t]{l}$+1?$\end{tabular}}}}%
    \put(0,0){\includegraphics[width=\unitlength,page=4]{rcount.pdf}}%
    \put(0.665,0.315){\color[rgb]{0,0,0}\makebox(0,0)[lt]{\lineheight{1.25}\smash{\begin{tabular}[t]{l}NSR\end{tabular}}}}%
    \put(0,0){\includegraphics[width=\unitlength,page=5]{rcount.pdf}}%
    \put(0.058,0.574){\color[rgb]{0,0,0}\makebox(0,0)[lt]{\lineheight{1.25}\smash{\begin{tabular}[t]{l}$c$\end{tabular}}}}%
    \put(0,0){\includegraphics[width=\unitlength,page=6]{rcount.pdf}}%
  \end{picture}%
\endgroup%

    \caption{Recurrent Counting with the NSR. The value $c$ stores the counter. The NSR compares the $i-$th element $x_i$ against the first element $x_0$. On equality the NSR can choose to control the conditional incrementer $+1?$ to increment the counter or to keep $c$'s value. For the MLP experiments, we replace the NSR through an MLP.}
    \label{fig_rcount}
\end{figure}

\section{Runtimes}
Thanks to the NSR's data efficiency, the training sets for most problems can be kept small. We can train all tasks but the shortest path problem and the MNIST task on CPUs in a matter of minutes. The shortest path model trains less than $2$ hours, then the bulk of computation is spent on the extrapolation tests. This is because the implementation scales quadratically with the number of nodes. Test times vary largely, depending on which number of nodes is sampled. In the worst case, we manage to complete training with all extrapolation tests in $18$ hours. Experiments are run as jobs on a cluster so we cannot report the exact CPU type--- it is either cores of a Dual Octa-Core Intel Xeon E5-2690 or a Dual Octa-Core Intel Xeon E5-2690 v2 CPU. For all fast experiments, we use $2$ CPU cores, while using $8$ cores for the shortest path computation. The MNIST model on a single NVIDIA Geforce Gtx Titan takes around $9$ hours to train for $20$ epochs, for both the NSR and MLP.

\newpage
\section{Scaling results for all comparisons}
\label{sec_allfloats}

\begin{minipage}{\textwidth}
    \centering
    \includegraphics[width=\textwidth]{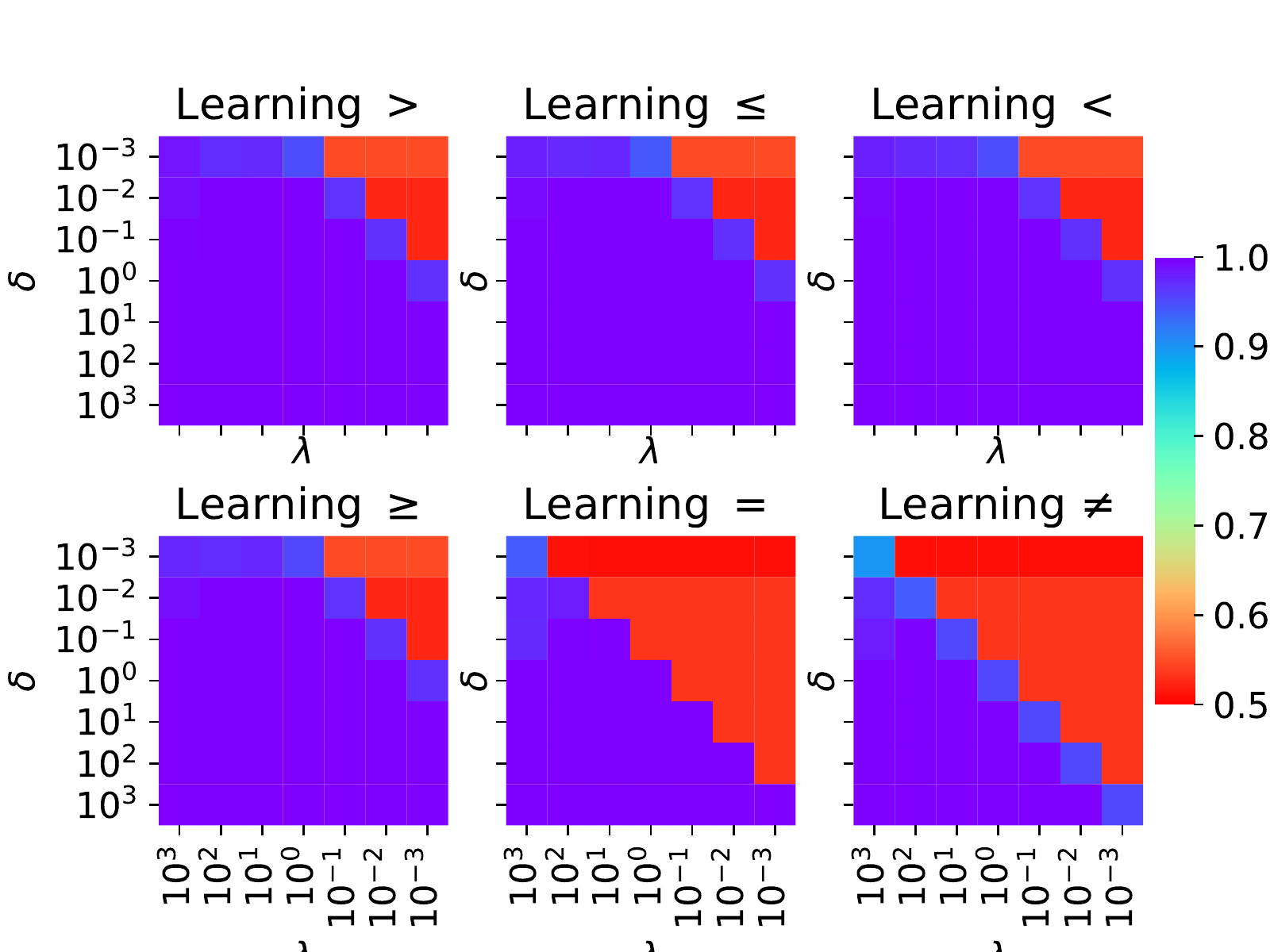}
    \captionof{figure}{Experiment for all comparison as in Figure~\ref{fig_floats} with identical setup. We can see there are four ``easy'' comparisons (also the ones the NSR solves perfectly in Table~\ref{tab_compresults}. The NSR is robust to changes in $\delta$ for the easy comparisons as long as $\lambda^{-1}$ is within $3$ orders of magnitude. Equalities $=$ and $\ne$ are more difficult where where $\lambda > \delta^{-1}$ is needed to train succesfully.}
    \label{fig:my_label}
\end{minipage}

\newpage~
\newpage

\section{Example weights for the NSR and MLP}
\begin{minipage}{0.99\textwidth}
  \captionof{table}{Weights to solve comparisons for the MLP network from Figure for integer inputs~\ref{fig_2nnlayout}. Inputs are vectors with two elements $x_1$ and $x_2$.}
  \label{tab_mlpweights}
  \centering
  \begin{tabular}{ccccccccccccc}
    \toprule
    Comparison & $V_{11}$ & $V_{12}$ & $V_{21}$ & $V_{22}$ & $b_{1}$ & $b_{2}$ & $H_1$ & $H_2$ & $b$\\\midrule
    $x_1>x_2$ & 100 & 0 & -100 & 0 & -50 & -50 & 100 & 0 & -50\\
    \midrule
    $x_1\leq x_2$ & 100 & 0 & -100 & 0 & 50 & -50 & -100 & 0 & 50\\
    \midrule
    $x_1<x_2$ & -100 & 0 & 100 & 0 & -50 & -50 & 100 & 0 & -50\\
    \midrule
    $x_1\geq x_2$ & -100 & 0 & 100 & 0 & -50 & -50 & -100 & 0 & 50\\
    \midrule
    $x_1= x_2$ & 100 & -100 & -100 & 100 & -50 & -50 & -100 & -100 & 50\\
    \midrule
    $x_1\ne x_2$ & 100 & -100 & -100 & 100 & -50 & -50 & 100 & 100 & -50\\
    \bottomrule
  \end{tabular}
\end{minipage}
\vspace{1cm}
\begin{minipage}{0.49\textwidth}
    \centering
    \includegraphics[width=0.8\textwidth]{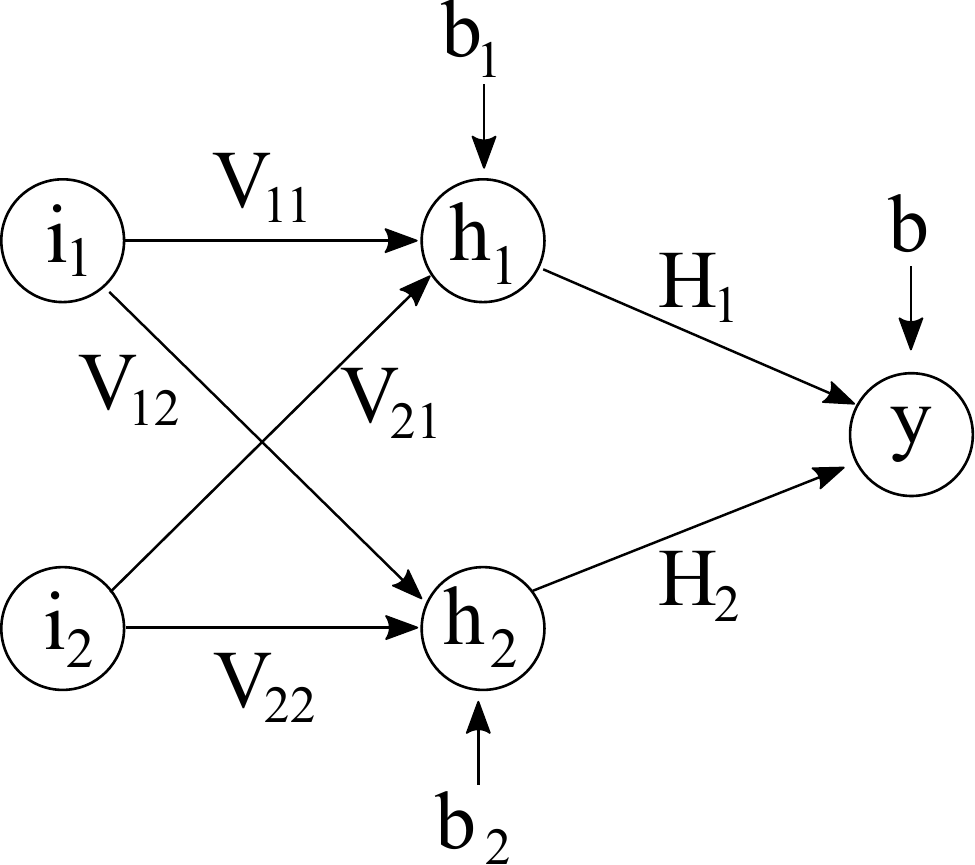}
    \captionof{figure}{Two-layer feedforward network with one hidden layer using sigmoid activation and an output layer using softmax activation.}
    \label{fig_2nnlayout}
\end{minipage}
\begin{minipage}{0.49\textwidth}
  \captionof{table}{Example values for the learnable weights of the NSR in Figure~\ref{fig_nsrarchitecture} to solve different comparisons for integer inputs. Inputs are vectors with two elements $x_1$ and $x_2$.}
  \label{tab_nsrweights}
  \centering
  \begin{tabular}{cccccccccc}
    \toprule
    Task & $V_{11}$ & $V_{12}$ & $V_{21}$ & $V_{22}$ & $\wsign$ & $\wzero$ & $b$ \\\midrule
    $x_1>x_2$ & 1 & 0 & 0 & 1 & 100 & 0 & -50\\
    \midrule
    $x_1\leq x_2$ & 1 & 0 & 0 & 1 & -100 & 0 & 50\\
    \midrule
    $x_1<x_2$ & 1 & 0 & 0 & 1 & -100 & -100 & 50\\
    \midrule
    $x_1\geq x_2$ & 1 & 0 & 0 & 1 & 100 & 100 & -50\\
    \midrule
    $x_1=x_2$ & 1 & 0 & 0 & 1 & 0 & 100 & -50\\
    \midrule
    $x_1\ne x_2$ & 1 & 0 & 0 & 1 & 0 & -100 & 50\\
    \bottomrule
  \end{tabular}
\end{minipage}

\end{document}